\documentclass[letterpaper]{article} 
\usepackage{aaai23}  
\usepackage{times}  
\usepackage{helvet}  
\usepackage{courier}  
\usepackage[hyphens]{url}  
\usepackage{graphicx} 
\usepackage{natbib}  
\usepackage{caption} 
\usepackage{algorithm}
\usepackage{diagbox}
\usepackage{algorithmic}
\usepackage{graphicx}
\usepackage{dblfloatfix}
\usepackage{textcomp}
\usepackage{xcolor}
\usepackage[inline]{enumitem}

\usepackage{newfloat}
\usepackage{listings}
\newcommand{\nick}[1]{{\textcolor{orange}{}}}
\newcommand{\ajitha}[1]{{\textcolor{red}{}}}
\newcommand{\jose} [1] {{\textcolor{blue}{}}}
\newcommand{\perry} [1] {{\textcolor{green}{}}}

\newcommand{\nicklouloudakis}[1]{{\textcolor{orange}{}}}
\newcommand{\jocare}[1]{{\textcolor{blue}{}}}

\usepackage{subcaption} 
\nocopyright

\title {Exploring Effects of Computational Parameter Changes \\ to Image Recognition Systems}
\author {
    Nikolaos Louloudakis\textsuperscript{\rm 1},
    Perry Gibson\textsuperscript{\rm 2}, 
    Jos\'e Cano\textsuperscript{\rm 2}, 
    Ajitha Rajan\textsuperscript{\rm 1}
}
\affiliations {
    \textsuperscript{\rm 1}The University of Edinburgh\\
    \textsuperscript{\rm 2}University of Glasgow\\
    n.louloudakis@ed.ac.uk, p.gibson.2@research.gla.ac.uk, jose.canoreyes@glasgow.ac.uk,
    arajan@ed.ac.uk
}

\begin{document}

\maketitle


\begin{abstract}

Image recognition tasks typically use deep learning and require enormous processing power, thus relying on hardware accelerators like GPUs and FPGAs for fast, timely processing. Failure in real-time image recognition tasks can occur due to incorrect mapping on hardware accelerators, which may lead to timing uncertainty and incorrect behavior.
Owing to the increased use of image recognition tasks in safety-critical applications like autonomous driving and medical imaging, it is imperative to assess their robustness to changes in the computational environment as parameters like deep learning frameworks, compiler optimizations for code generation, and hardware devices are not regulated with varying impact on model performance and correctness. 
In this paper we conduct robustness analysis of four popular image recognition models (MobileNetV2, ResNet101V2, DenseNet121 and InceptionV3) with the ImageNet dataset, assessing the impact of the following parameters in the model's computational environment: (1) deep learning frameworks; (2) compiler optimizations; and (3) hardware devices. 
We report sensitivity of model performance in terms of output label and inference time for changes in each of these environment parameters. 
We find that output label predictions for all four models are sensitive to choice of deep learning framework (by up to 57\%) and insensitive to other parameters.
On the other hand, model inference time was affected by all environment parameters with changes in hardware device having the most effect. The extent of effect was not uniform across models. 

\end{abstract}


\section{Introduction}

The first step in achieving environmental
perception in autonomous vehicles (AV) is to detect objects using object detection algorithms that is central for recognizing and
localizing objects such as pedestrians, traffic lights/signs, other
vehicles, and barriers in the AV vicinity. 
Typically object detection algorithms use Deep Neural Networks (DNNs) for image recognition and localisation, as they can learn and extract more complex features.

\begin{figure}[!t]
 \centering
 \includegraphics[width=0.97\columnwidth]{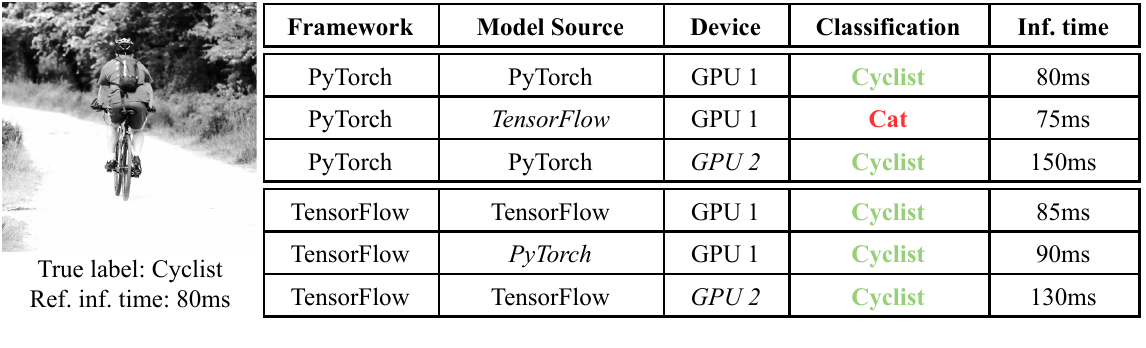}
 \caption{\label{fig:motivation} Example showing how varying factors can impact the classification accuracy of an image recognition model.}
\end{figure}

Much of the existing literature for assessing robustness and safety of image recognition has focused on testing the DNN structure and addressing bias in the training dataset through adversarial testing and data augmentation~\cite{zhang2018deeproad, tian2018deeptest, guoCoverageGuidedDifferential2021}.
Existing techniques have failed to consider safety violations caused by interactions of the DNN with the underlying computational environment: both software and hardware.
This can include the Deep Learning (DL) frameworks (e.g., TensorFlow, PyTorch, etc), compiler optimizations for device code generation (e.g., operator fusion, loop unrolling), and the hardware accelerators they run on (e.g.GPUs).
As an illustrative example, consider an image recognition model to classify the image in Figure~\ref{fig:motivation} with a true label ``Cyclist'' and reference inference times of $80$ms with PyTorch and $85$ms with TensorFlow.
The figure illustrates how changing the DL framework or device (italicized in the table) can impact the output label and/or model inference time.

In this paper, we conduct an empirical study to evaluate the robustness of image recognition models in the presence of changes in the computational environment.
We consider the following parameters in the computational environment: (1) Source DL framework for the DNN model; (2) Compiler optimizations; and finally (3) GPU accelerator devices. 
We assess the robustness of four widely used image recognition models with respect to the output label and inference time when changing each of these environment parameters. 
It is important to check changes in the output label, as it directly affects model correctness. On the other hand, inference time is an important consideration for timing safety in real-time perception systems within applications like self-driving cars where there is a performance requirement for object detection models to return results within a fixed time~\cite{dreossi2019verifai}. 

Overall, we find that varying DL frameworks significantly impacts output label and inference time of the model. Varying hardware accelerators and compiler optimization do not affect model output but have a significant effect on inference time. In summary, we make the following contributions: 
\begin{enumerate}
    \item Assess robustness of image recognition model \emph{outputs} with respect to changes in the computational environment: DL frameworks, compiler optimizations, and hardware accelerators;
    
    \item Assess robustness of model \emph{inference time} with respect to changes in the computational environment: DL frameworks, compiler optimizations, and hardware accelerators;
\end{enumerate}

\section{Background}

Figure~\ref{fig:systems_stack} gives an overview of the typical layers in the deep learning systems stack~\cite{iiswc_2018}.
Much of the existing work has focused on testing and robustness with respect to the top two layers, \emph{Datasets} and \emph{Models}. 
In this paper, we consider robustness with respect to the bottom three layers that make up the computational environment 
required for executing a given DL model, that includes the deep learning framework, related systems software, and the underlying hardware. 
We describe the relevant parts of the computational environment in the sections below. 
In addition, we provide a brief overview of image recognition models that is the focus of this study. 

\subsection{Deep Learning Frameworks}
\begin{figure}[!t]
 \centering
 \includegraphics[width=0.99\columnwidth]{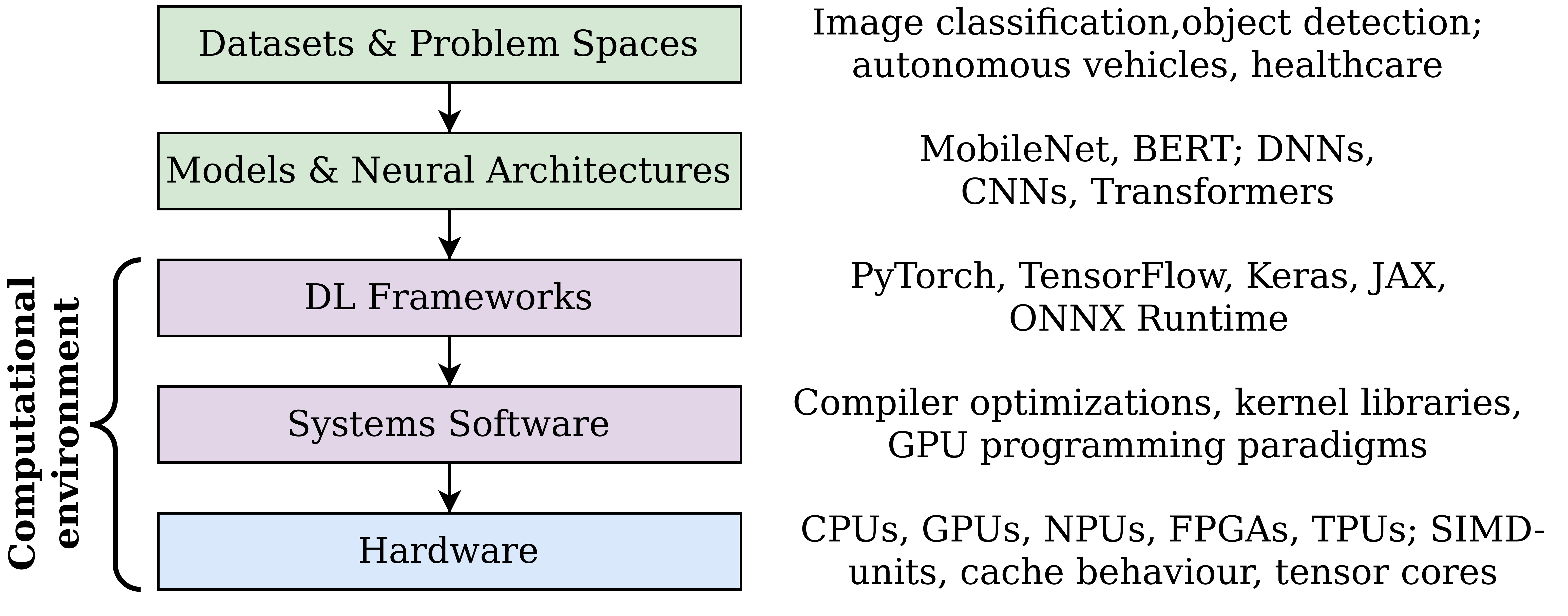}
 \caption{\label{fig:systems_stack} Relevant layers in the deep learning systems stack.}
\end{figure}

Deep Learning Frameworks, shown as the third layer in Figure~\ref{fig:systems_stack}, provide utilities such as model declaration, training and inference to machine learning engineers. 
For our study, we use four DL frameworks that are widely used in the community: Keras, PyTorch, TensorFlow (TF), and TensorFlow Lite (also known as TFLite).
We use these frameworks as sources for the image recognition models, as each has its own native definition for the models.
We briefly describe each of the four frameworks below.

\textbf{Keras}~\cite{chollet2015keras} is a high-level DL framework, providing APIs for effective deep learning usage. Keras acts as an interface for TensorFlow, and we aim to observe potential overheads and bug introductions from the extra layer of complexity.

\textbf{PyTorch}~\cite{pytorch} is an open source machine learning framework based on the Torch library and developed by Meta AI team. It supports hardware acceleration for tensor computing operations.
    
\textbf{TensorFlow (TF)}~\cite{tensorflow2015-whitepaper} is an open-source DL framework, developed by Google, and widely used for training and inference of DNNs.

\textbf{TensorFlow Lite (TFLite)}~\cite{tensorflow2015-whitepaper} is a lightweight version of TensorFlow and part of the original TensorFlow library, providing framework focused only on the inference of neural networks on mobile and lightweight devices.


\subsubsection{Framework Conversion}

Conversion of models between DL frameworks can be a complex task, and thus many frameworks redefine common DNN architectures natively, and often training said model from scratch.
We refer to such models as a \emph{``native model''}, and models that have been converted to another DL framework as a \emph{``converted model''}.
Systems such as ONNX~\cite{onnx} and MMdnn~\cite{liuEnhancingInteroperabilityDeep2020} attempt to provide common intermediate formats for translation between DL frameworks, however these processes can still be error prone, and have issues around support for bespoke operators.
Note that the DL framework used to design and train a model may not be the same as that used to deploy the model.
Hence, it is worth exploring potential errors that may be introduced during framework translation.
We only consider two framework translations in our experiments: (1) TF to TFLite; and (2) PyTorch to TFLite.
This is because TFLite is a deployment-only framework and thus is more likely to be the final software environment that developers convert their models to.

We will explore other framework conversions in our future work. 


\subsection{Systems Software: Apache TVM}

\begin{figure}[!t]
 \centering
 \includegraphics[width=0.99\columnwidth]{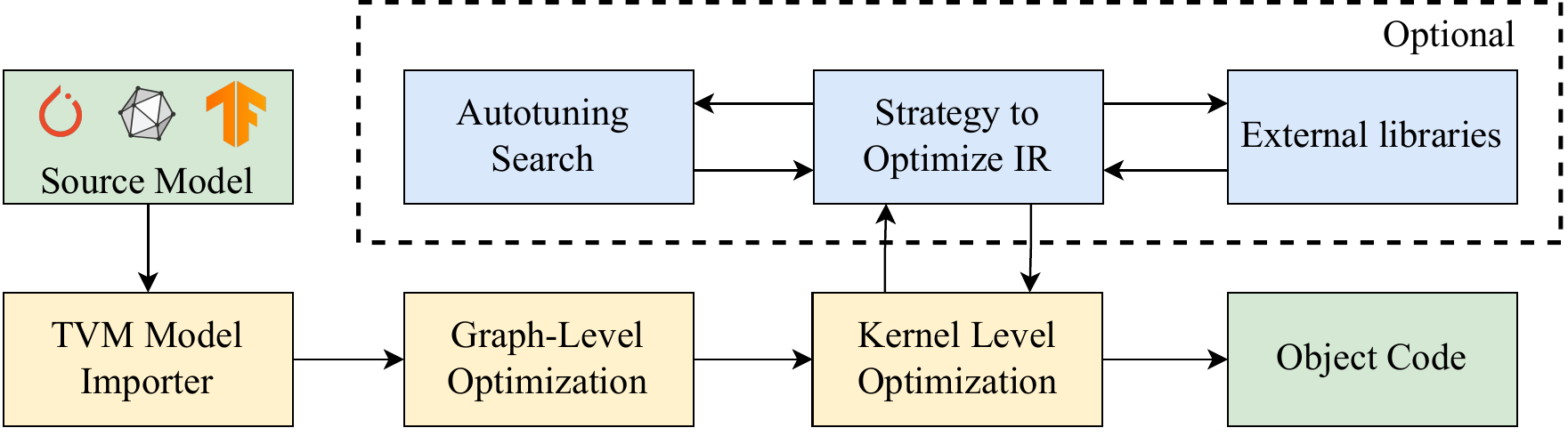}
 \caption{\label{fig:tvm_simple} Overview of DNN compilation in Apache TVM.}
\end{figure}

Apache TVM~\cite{tvm} is an end-to-end machine learning compiler framework for CPUs, GPUs, and accelerators.
It generates optimized code for specific DNN models and hardware backends, allows us to import DNN models from a range of DL frameworks, and provides profiling utilities such as per-layer inference times.
A simplified representation of Apache TVM can be seen in Figure~\ref{fig:tvm_simple}.
TVM's support of several DL frameworks, optimization settings, and hardware accelerators made it a suitable choice explore varying different environment parameters in our experiments.
TVM provides direct importers for models from most popular DL frameworks, which load said models as a TVM computation graph.

The first level of optimization available in TVM is graph-level optimizations, which is the focus of this study.
These optimizations impact the full model and include operator fusion (e.g., batch normalization, activation functions), elimination of common subexpressions, and potentially unsafe optimizations such as fast math. 

TVM also supports optimizations for a given operation type (e.g., convolutional layers, matrix-multiplications) such as loop tiling, loop re-ordering, unrolling, vectorization, auto-tuning~\cite{chenLearningOptimizeTensor2018}, and auto-scheduling~\cite{zhengAnsorGeneratingHighPerformance2020}, among others.

TVM also supports third-party libraries such as cuDNN~\cite{chetlur2014cudnn} and the Arm Compute Library~\cite{ComputeLibrary2022}.

\subsection{The Perception AI Models}

A common benchmark for Perception AI models is the ImageNet image classification dataset~\cite{ILSVRC17}, which requires assigning one of a possible 1000 class labels to RGB images of size $224\times224$ pixels.
For solving Perception AI problems, such as classification and semantic segmentation, convolutional neural networks (CNNs) are commonly used, which are DNNs characterized by convolutional layers.
Transformers-based architectures~\cite{vaswani2017attention} have begun to provide competitive results in the past two years~\cite{daiCoAtNetMarryingConvolution2021,zhaiScalingVisionTransformers2022}, however are still maturing.
Thus for our evaluation we explore four widely used CNN models: MobileNetV2~\cite{mobilenetv2}, ResNet101V2~\cite{resnet}, DenseNet121~\cite{densenet}, and InceptionV3~\cite{inceptionv3}.
These models are widely known and extensively used for classification and semantic segmentation operations, as well as being the ``backbone network'' for other tasks such as object detection~\cite{chiuMobilenetSSDv2ImprovedObject2020}.

All four models have native definitions within the DL frameworks under study.

\section{Related Work}

Existing work has primarily focused on robustness of the dataset and model architecture layers (top two layers), shown  in Figure~\ref{fig:systems_stack}. DeepXPlore~\cite{deepxplore} applies whitebox testing, by measuring neuron coverage, identifying similar DNNs for cross-reference and generating adversarial inputs to detect faults. This work has been extended by DLFuzz that attempts to minutely mutate inputs to improve neuron coverage~\cite{guoCoverageGuidedDifferential2021}. DeepHunter~\cite{xie2019deephunter} applies fuzzing (i.e., generation of random, invalid and unexpected inputs) to DNNs, aiming to maximize coverage of the system and potentially discover faults. DeepTest~\cite{tian2018deeptest}, a tool that modifies images using linear \& affine transformations, generates inputs simulating different weather conditions and real-world phenomena to test the robustness and validity of DNNs to changing weather conditions in autonomous driving. DeepRoad~\cite{zhang2018deeproad} uses GAN-based metamorphic testing to generate inputs that simulate extreme weather conditions, such as heavy rain and snow. DeepBillboard~\cite{zhou2020deepbillboard} explores the potential of physical world adversarial testing utilizing Billboard inputs. For a more comprehensive overview of adverarial examples for images, we refer the readers to a survey~\cite{shorten2019survey}. 

Robustness with respect to layers in the computational environment, seen in Figure~\ref{fig:systems_stack}, has received little attention. 

With respect to the DL Frameworks layer, some attempts have been made to explore the effect of DL frameworks towards model performance. In particular, some benchmarking analysis has been conducted towards training and inference time analysis and performance~\cite{benchmarkdlswtools,benchmarkingdlframeworkmetrics,dlbench}. In addition, a survey~\cite{studydlframeworks} explores various parameters and their effect towards model accuracy and execution time. However, both contributions utilize experiment sets of limited of model numbers, DL frameworks, input dataset and variety of hardware acceleration devices, providing useful results but in a small scale. Our contribution aims to extend this work against real-world, challenging conditions and scenarios, exploring the effects of a challenging dataset, plus a wide variety of models and hardware acceleration devices capabilities, a setup much closer to real-world environments of safety-critical systems.

In addition, CRADLE~\cite{pham2019cradle} attempts to detect and localise inconsistencies between models sourced from different DL frameworks by comparing their outputs and analysing model execution. 
LEMON~\cite{wang2020lemon} is a framework that generates model mutations to detect discrepancies in DL frameworks used in Neural Networks. Both CRADLE and LEMON aim to detect faults in DL frameworks by comparing it to other frameworks. Impact of changing DL framework on model performance is not considered in these paper and that is the focus in our work. 

For the systems software layer in Figure~\ref{fig:systems_stack}, a recent study~\cite{dlcompileroptimisationbugs} examined bugs introduced by different deep learning compilers. Incorrect optimization code logic accounted for 9\% of the bugs introduced by compilers. Other compiler bugs presented in the study include misconfiguration, type problem, API misuse, incorrect exception handling, incompatibility. In our study, we examine the effect of changing compiler optimisations on model performance. We will examine the effect of other compiler bugs in our future work. 

Finally, for the hardware layer in Figure~\ref{fig:systems_stack},~\cite{humbatova2020taxonomy} created a taxonomy of faults encountered in DNNs used in object Detection. The authors surveyed commits, issues and pull requests from 564 GitHub projects and 9,935 posts from Stack Overflow and interviewed 20 researchers and practitioners. The study revealed \textit{GPU
related} bugs to be one of the five main categories faults in deep learning tasks like object detection. The study, however, did not explore the impact of these bugs on model performance.  The other four categories of faults were \textit{API}, \textit{Model}, \textit{Tensors and Inputs} and \textit{Training} that relate to the top two layers in Figure~\ref{fig:systems_stack}. In this paper, we are primarily interested in the effect of changes in the bottom three layers of the stack that make up the computational environment on model performance, in terms of output and model inference time. This has not been systematically explored in the literature.

\section{Experiments}

\begin{table}[]

\caption{Accuracy of native models on the ImageNet dataset.}
\label{tab:native_accuracy}
\centering
\fontsize{7}{9}\selectfont
\begin{tabular}{|l|c|c|c|c|}
\hline
\backslashbox{\textbf{DNN Model}}{\textbf{Framework}} & \rotatebox[origin=c]{90}{~PyTorch~} & \rotatebox[origin=c]{90}{Keras} & \rotatebox[origin=c]{90}{TF} & \rotatebox[origin=c]{90}{TFLite} \\\hline
   ResNet101      & 81.9      & 76.4       & 77.0        &   77.0    \\\hline
InceptionV3          &  77.3      & 77.9        &   78.0         &   78.0     \\\hline
MobileNetV2       &   72.2        & 71.3           &   71.9          &   71.9          \\\hline
DenseNet121 &    74.4         &     75.0         &     N/A        &    N/A    \\\hline
\end{tabular}
\end{table}

We consider four different CNN (image recognition) models: MobileNetV2, ResNet101V2, DenseNet121, and InceptionV3.
Each model is evaluated in the TVM compiler framework (v0.8.0), imported via ONNX~\cite{onnx}, with all possible combinations of values for each of the following environment parameters:
\begin{description}
    \item[DL Frameworks:] We consider Keras, TF, TFLite, and PyTorch as sources for our models.
    We selected these frameworks as they are widely used in the deep learning community~\cite{SurveyCNNs}. 
    The accuracy of the native version of each model is shown in Table~\ref{tab:native_accuracy}.
    We verify that each model in correctly imported into TVM by comparing the output labels with the source framework.
    \item[Framework Conversion:] We examine the effect of framework conversion on model robustness.
    In particular, we convert native models from PyTorch and TF to TFLite models, and examine labelling differences introduced by the conversion.

    \item[Compiler Optimizations:]
    We explored the impact of different levels of TVM graph-level compiler optimization: basic, default, and extended variants.
    
    \textbf{Basic} (\texttt{o0}) applies only ``inference simplification'', which generates simplified expressions with the same semantic equivalence as the original DNN.

    \textbf{Default} (\texttt{o2}) applies all optimizations of \texttt{o0}, and in addition fusion of operators such as ReLU activation functions, as well as constant folding.

    \textbf{Extended} (\texttt{o4}) applies all optimizations from Default and a number of additional ones. 
    For example: enabling ``fast math'' (which allows the compiler to  break strict IEEE standard compliance for float operations if it could improve performance), allowing modification of data layouts, and eliminating subexpressions with multiple occurrences.
    \item[Devices:] We used four different hardware devices, featuring GPU accelerators of varying capabilities. 
    More specifically, we used:
    \begin{description}
        \item[\textbf{Server:}] Intel-based server featuring an Nvidia Tesla K40c (GK11BGL) GPU; 
        \item[\textbf{Xavier:}] Nvidia AGX Xavier featuring an Nvidia Volta GPU;
        \item[\textbf{Local:}] Laptop featuring an Intel(R) GEN9 HD Graphics NEO;
        \item[\textbf{Hikey:}] Hikey 970 board featuring an Arm Mali-G72 GPU;
    \end{description}
\end{description}
\noindent Note that \texttt{Server} represents a high-end Nvidia GPU, \texttt{Xavier} a mid-end Nvidia GPU, \texttt{Local} a low-end Intel GPU, and \texttt{Hikey} a mobile-class Arm GPU.
For the Nvidia devices we use TVM to generate CUDA code, and for the Hikey and Intel GPUs we generate OpenCL code.

In total, we evaluate $276$ model variants from $4$ Models $*\ (4$ DL Frameworks $+\ 2\ $DL Framework Conversions) $*\ 4$ Devices $*\ 3$ Optimizations $-\ (12$ Keras native models$))$.
We subtract $12$ Keras native models, since we were unable to compile the InceptionV3 model on any device or under any optimization setting.  
We discuss challenges faced in compiling and executing certain configurations in the Execution Issues Section~\ref{subsec:execution_issues}. 


\subsubsection{Dataset}

We use the ImageNet Large Scale Visual Recognition Challenge 2017 (ILSVRC2017)~\cite{ILSVRC17} image classification test dataset for our experiments, consisting of 5500 RGB images, of size $224\times224$ pixels.
The task is to produce an output label classifying the image, out of 1000 possible labels (all our models were pre-trained on the ImageNet dataset, see Table~\ref{tab:native_accuracy} for their accuracy).


\subsection{Robustness Measurements}

Our experiments are aimed at evaluating (1) Robustness of Model Output, and (2) Robustness of Model Execution Time. 
We describe these measurements below. 

\subsubsection{1. Robustness of Model Output}


For every image input in the ImageNet~\cite{ILSVRC17} test dataset, we record the top-ranked output label for every combination of environment parameters. 
We then conduct pairwise comparison of the labels for the same image input while varying each environment parameter in turn. 
For instance, over a single image, we would compare the output label from InceptionV3 from Keras against that of PyTorch while keeping the device and compiler optimization constant.  
We then compute total dissimilarity in labels across all images in the dataset for every pairwise environment parameter variation.


\subsubsection{2. Robustness of Model Execution Time}

We use the term \emph{model execution time} to mean model inference time which is the processing time of the network model without including the time to load images and any pre-processing that may be needed. 
We record model execution time for every image in the dataset and with every model configuration. We repeated executions 10 times and recorded average time per image. 
We compare execution times for model configurations across all images in the dataset using box plots and mean execution time. 


\subsection{Execution Issues}
\label{subsec:execution_issues}

All environment parameter combinations could not be executed with all models due to the following incompatibility issues: \\
    \noindent 1. The DenseNet121 model from TF and TFLite resulted in incorrect output labels for most images. 
    The output labels remained constant regardless of image, even when running within TensorFlow itself.
    Although we sourced the model from the TensorFlow website, we believe that the model has been deprecated, and thus does not behave as expected.
    This is because it no longer appears in the list of pre-trained models within the TensoFlow repository\footnote{\url{https://github.com/tensorflow/models/tree/master/research/slim#pre-trained-models}}. 
    \noindent 2. InceptionV3 sourced from Keras was problematic, in the sense that we were unable to succesfully import it into TVM.
    We attempted using TVM's Keras model importer, as well as importing the model via ONNX, however in both cases the import failed.
    We did not experience this issue with any other version of InceptionV3.
    
    \noindent 3. For ResNet101 sourced  from PyTorch, we selected the V1 version the model instead of V2 as the V2 version was not provided in the official PyTorch repository.
    The version difference may have a larger effect on model inference time when we compare across DL frameworks.
    We therefore ask the readers to take this into consideration for results involving ResNet101 sensitivity to DL framework.
    
    \noindent 4. Regarding MobileNetV2, we experienced problems when executing it on the \texttt{Xavier} device, as we received a \texttt{CUDA\_ERROR\_INVALID\_PTX} error. We do not consider this device configuration for MobileNetV2 in our experiments.



\section{Results}

For each image recognition model, we discuss robustness of (1) output label prediction and (2) model inference time in the presence of changes in the the DL framework from which models are sourced, compiler optimizations, and hardware devices. 


\subsection{Robustness of Model Output (1)}
We vary one environment parameter at a time (while fixing the others) -- the DL framework, compiler optimization level, GPU device -- and examine their impact on output label prediction.
Table~\ref{tab:native_accuracy} shows the accuracy of native models on the ImageNet test dataset.
To take InceptionV3 as an example, all frameworks get approximately 78\% accuracy, however this does not mean that thy will be correct for the same 78\%.
Thus in the worst case, we would expect two frameworks would only agree on 56\% of labels (i.e., 44\% dissimilarity).


\subsubsection{Varying Deep Learning framework} 

\begin{figure*}[t]
\centering
    \begin{subfigure}{0.24\textwidth}
        \centering
        \includegraphics[width=\textwidth]{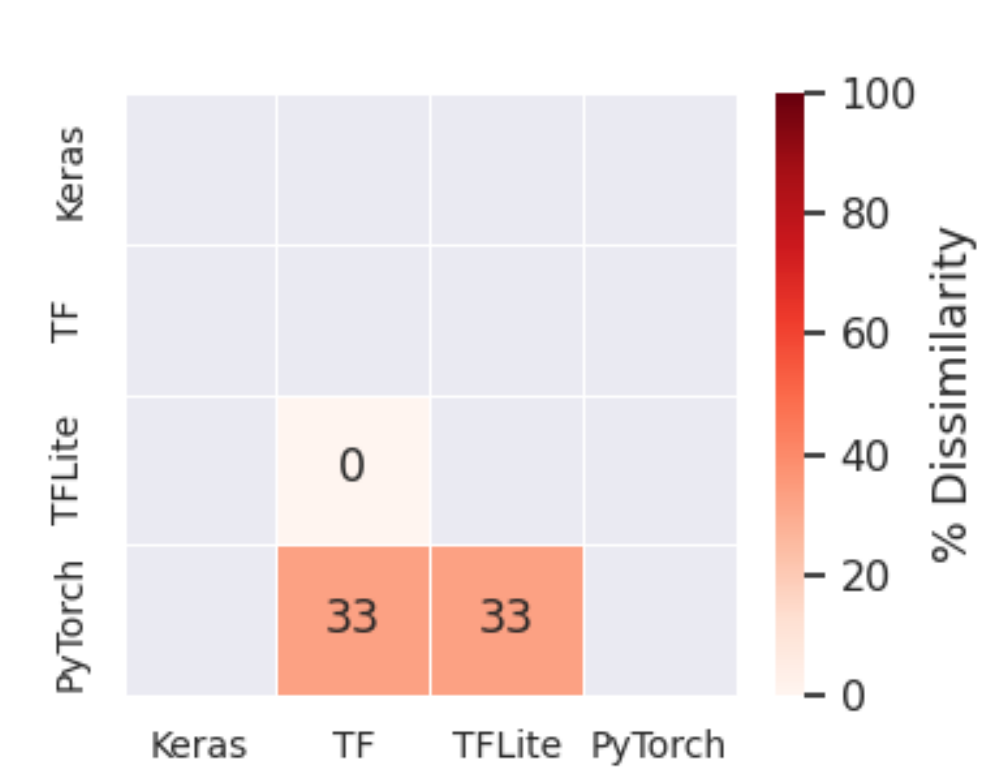}
 \caption{InceptionV3}
 \label{fig:inception}
 \end{subfigure}%
  \hspace*{\fill}   
  \centering
    \begin{subfigure}{0.24\textwidth}
    \centering
        \includegraphics[width=\textwidth]{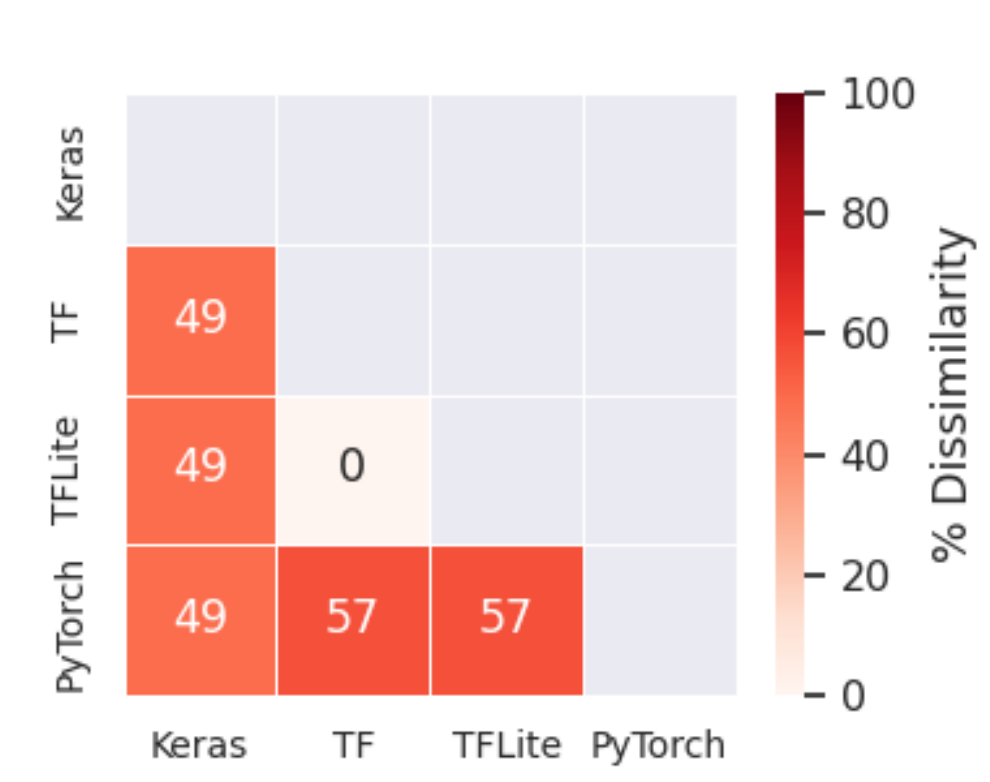}
 \caption{MobileNetV2}
 \label{fig:mobilenet}
    \end{subfigure}
    \begin{subfigure}{0.24\textwidth}
        \centering
        \includegraphics[width=\textwidth]{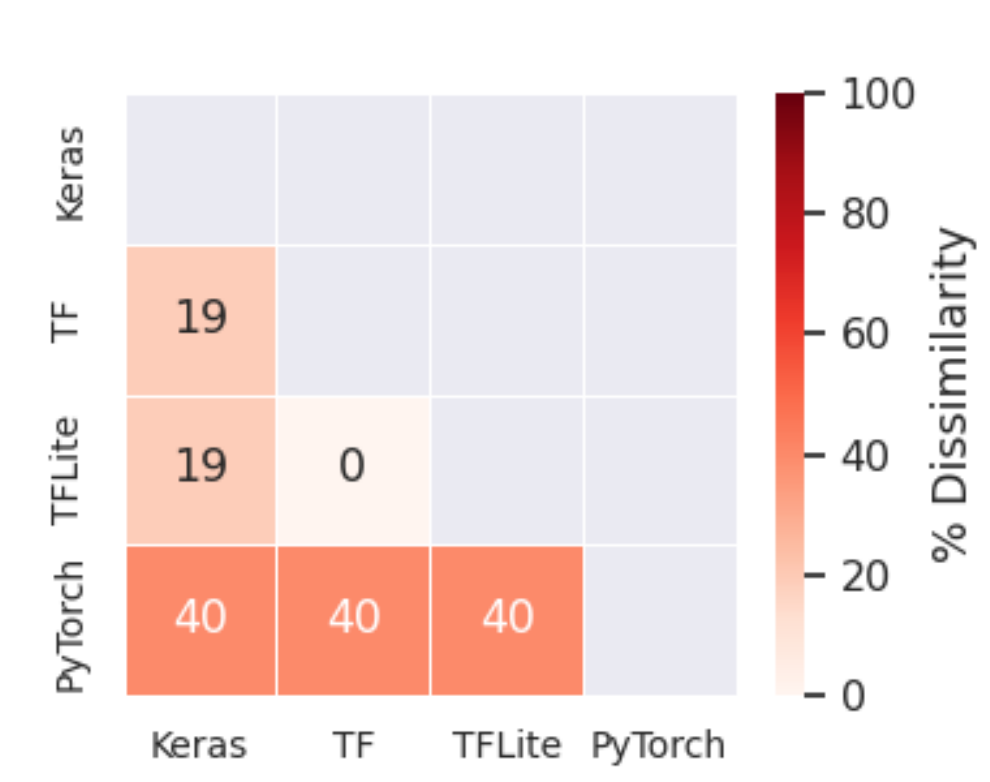}
 \caption{ResNet101}
 \label{fig:resnet}
 \end{subfigure}%
  \hspace*{\fill}   
  \centering
    \begin{subfigure}{0.24\textwidth}
    \centering
        \includegraphics[width=\textwidth]{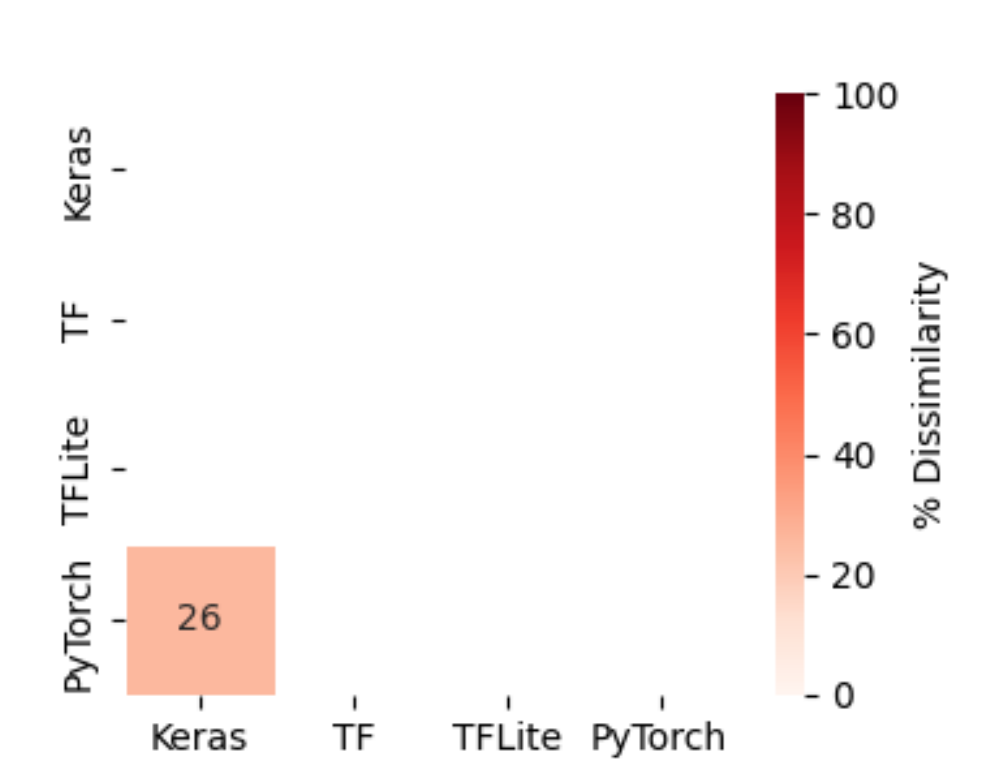}
 \caption{DenseNet121}
 \label{fig:densenet}
    \end{subfigure}%
 \caption{Pairwise comparison of output label dissimilarities (\%) between DL frameworks for our 4 models, running on \texttt{Server}, with \emph{Default} optimization level.}
  \label{fig:label_dissimilar}
\end{figure*}

We present results for the four models under study in Figure~\ref{fig:label_dissimilar}, with the TVM compiler optimization level set to \emph{Default} (\texttt{-o2}), and the hardware acceleration device set to \texttt{Server}.

We then vary the DL frameworks one at a time to compute sensitivity of model output label to that framework. 
Figures~\ref{fig:inception}--\ref{fig:densenet} show that models are acutely sensitive to the DL framework they are sourced from. 
Changes in the framework has a significant impact on output label, with MobileNetV2 exhibiting most discrepancy in output labels, in the range of $49-57\%$ for different DL frameworks.
We analyse each model below:

\begin{description}
    \item[InceptionV3] (Figure~\ref{fig:inception}) we observe a 33\% discrepancy between PyTorch versus both TF and TFLite. 
    No discrepancies were observed for TF versus TFLite; the same was true with other models.
    
    \item[MobileNetV2] (Figure~\ref{fig:mobilenet}) we observe a 49\% dissimilarity between Keras versus the other three frameworks and a 57\% dissimilarity for PyTorch versus TF and TFLite.
    We hypothesize that lower complexity and size of MobileNetV2 makes it less robust to changes in the framework.
    
    \item[ResNet101V2] (Figure~\ref{fig:resnet}) Keras has a 19\% dissimilarity against TF and TFLite. 
    PyTorch resulted in a 40\% dissimilarity against all the other frameworks. 

    \item[DenseNet121] (Figure~\ref{fig:densenet}) we find a 26\% dissimilarity for PyTorch versus Keras. 
    Owing to the execution issues with TF/TFLite (discussed in Execution Issues section), we do not report dissimilarities when changing to these frameworks.
\end{description}

Overall, TF and TFLite models always produce the same output, which suggests that the official TFLite models are successfully converted TF models, and contain the same parameters.
This is further confirmed by their achieving the same accuracy as seen in Table~\ref{tab:native_accuracy}.
However, when comparing against Keras and PyTorch, we observe significant differences. 
The extent of difference varies widely among the models, with MobileNetV2 being most sensitive to DL framework changes (see Figures~\ref{fig:mobilenet}).

\begin{figure}[!t]
 \centering
 \includegraphics[width=\columnwidth]{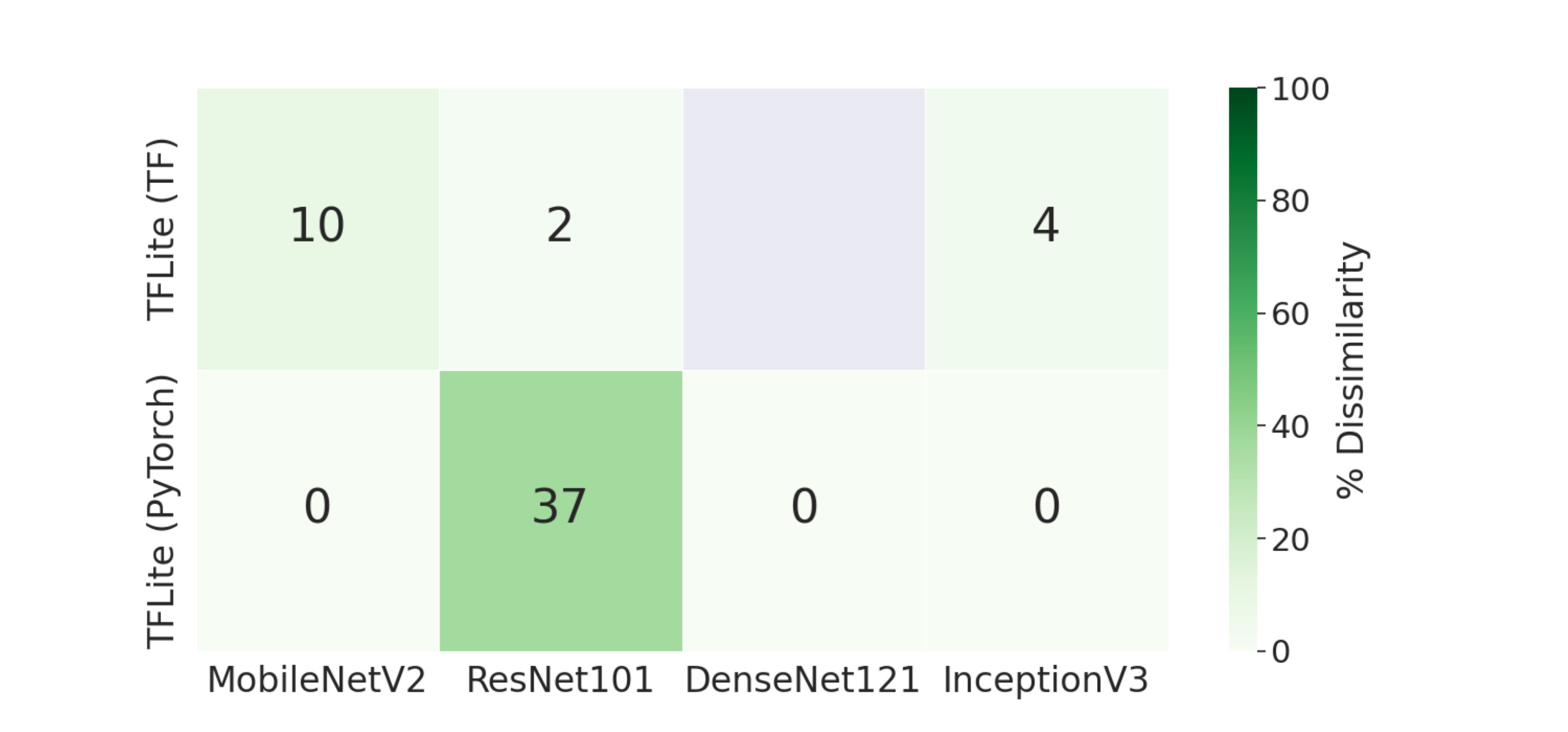}
 \caption{DL Framework Conversions Dissimilarities (\%).}
 \label{fig:conversions}
\end{figure}

\paragraph{Impact of DL Framework Conversion}
We explore the impact of framework conversion on model output by converting models from PyTorch and TF to TFLite, and evaluating the converted models in TVM.
For TF-to-TFLite, we were able to convert directly, whereas for PyTorch we had to convert to ONNX, then to TF, then to TFLite.
We then compare the output labels between the native and converted models, i.e. if any errors were introduced by converting the model.
The results are presented in Figure~\ref{fig:conversions}.
We observed 37\% discrepancies in output label when converting the ResNet101 model from PyTorch to TFLite, meaning that, for models converted from TF to TFLite, discrepancies are fewer and are in the range of 2\% (ResNet101) to 10\% (MobileNetV2). 

\paragraph{Analysis of Output Label Sensitivity}
Small changes in output labels when varying DL frameworks might be expected, since each framework has a specific implementation of the model and has been trained independently.

However, the extent of change in output labels (0-57\%) across images in the test dataset observed in all 4 models when changing DL frameworks is surprising. 
To gain a better insight, for 10 images from the test dataset producing different labels, we took a closer look at the convolutional layers within one of the models, DenseNet121, sourced from different frameworks. 
We compared layer activation tensors between the models to identify the layers involved in the label discrepancy.
We use an error threshold for comparing elements within the tensors.

\begin{figure}[t]
\centering
    \begin{subfigure}{0.49\linewidth}
        \centering
        \includegraphics[width=\textwidth]{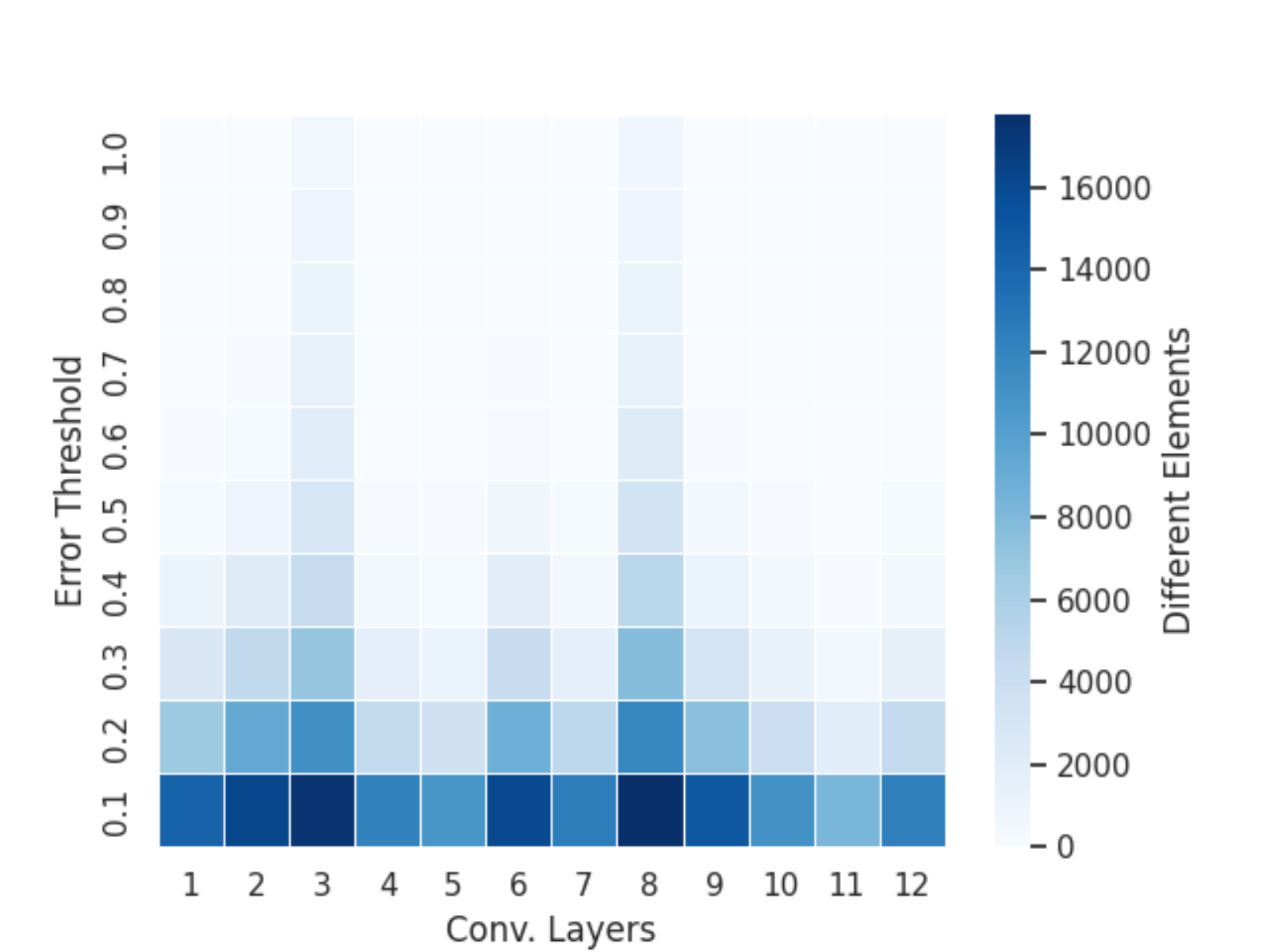}
 \caption{ImageNet 00002430}
 \label{fig:conv-activations1}
 \end{subfigure}%
  \hspace*{\fill}   
  \centering
    \begin{subfigure}{0.49\linewidth}
    \centering
        \includegraphics[width=\textwidth]{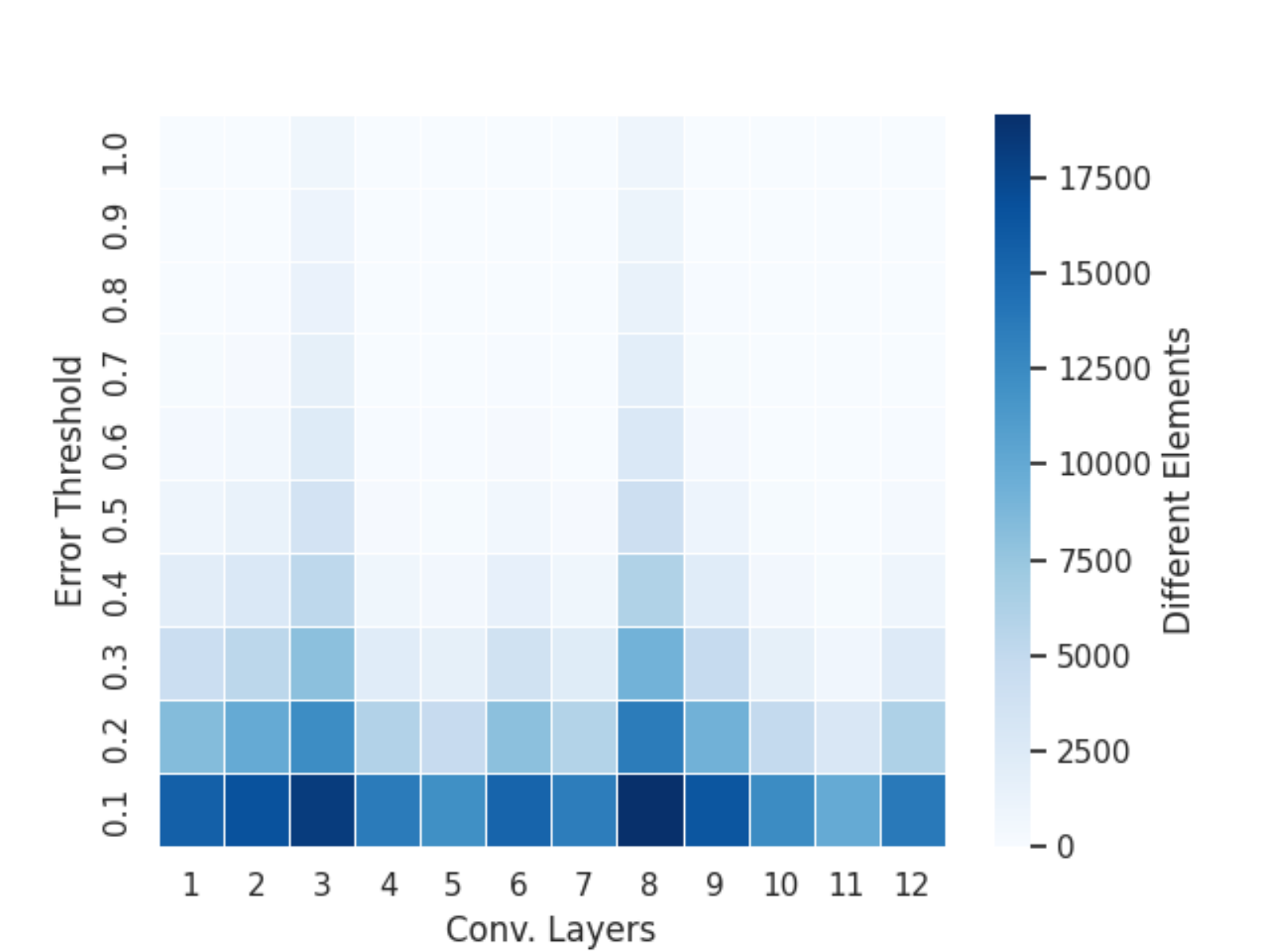}
 \caption{ImageNet 00004726}
 \label{fig:conv-activations2}
    \end{subfigure}
 \caption{Convolution activations comparison for DenseNet121.}
  \label{fig:conv-activations}
  \vspace{-10pt}
\end{figure}

Figures~\ref{fig:conv-activations1} and \ref{fig:conv-activations2} show the number of tensor elements that are different between corresponding layer activations\footnote{We show a subset of 12 convolutional layers from the model.} in DenseNet121 sourced from Keras, versus PyTorch for two images that produce different labels on them. 
We also show the effect of choosing different error threshold values.
We find for both images with label discrepancies that layers $3$ and $8$, followed by layers $2$, $9$ and $6$, have the highest number of differing tensor elements between the DenseNet121 variants.
A similar trend was found for most of the other 8 images with discrepancies that we investigated.
For the DenseNet121 model, Keras versus PyTorch, we find the aforementioned layers are worth investigating further for identifying source of model output sensitivity.

\subsubsection{Varying Compiler Optimizations}
We varied the optimization level within the TVM framework between \emph{Basic}, \emph{Default}, and \emph{Extended}. 
We kept the framework and device constant for assessing sensitivity to optimization level.

We found that varying compiler optimization levels causes no discrepancies in output labels for all four models. 

The lack of discrepancies/sensitivity is notable, since the \emph{Extended} (\texttt{-o4}) level enables unsafe math optimization that allows code violating IEEE \texttt{float} conventions to be generated.
Note that these potential unsafe perturbations were minor enough that all four models were resilient to them.
It is, however, worth considering robustness checks with respect to optimization levels in safety-critical domains, in case unsafe optimizations result in undesirable model outputs.

\subsubsection{Varying Hardware Accelerator}

With a fixed DL framework and compiler optimization (\texttt{-o2}), we compiled each of the four image recognition models on the four hardware devices: \texttt{Server}, \texttt{Xavier}, \texttt{Hikey}, and \texttt{Local}.

We found output label prediction for all four models with the ImageNet dataset was unaffected by changes in the hardware device and programming paradigm (OpenCL for Intel/Arm devices, and CUDA for Nvidia devices). 
This demonstrates that label predictions in our experiment are robust to device changes, at least when using Apache TVM as the code generator.
In our future work, we plan to explore the impact of varying the backend library, for example cuDNN~\cite{chetlur2014cudnn} and the Arm Compute Library~\cite{ComputeLibrary2022}.


\subsection{2. Robustness of Model Inference Time}
We vary one environment parameter at a time while fixing the others to check their impact on model inference time. 


\subsubsection{Varying Deep Learning Framework}

\begin{figure}[!t]
	\centering
	\includegraphics[width=0.7\columnwidth]{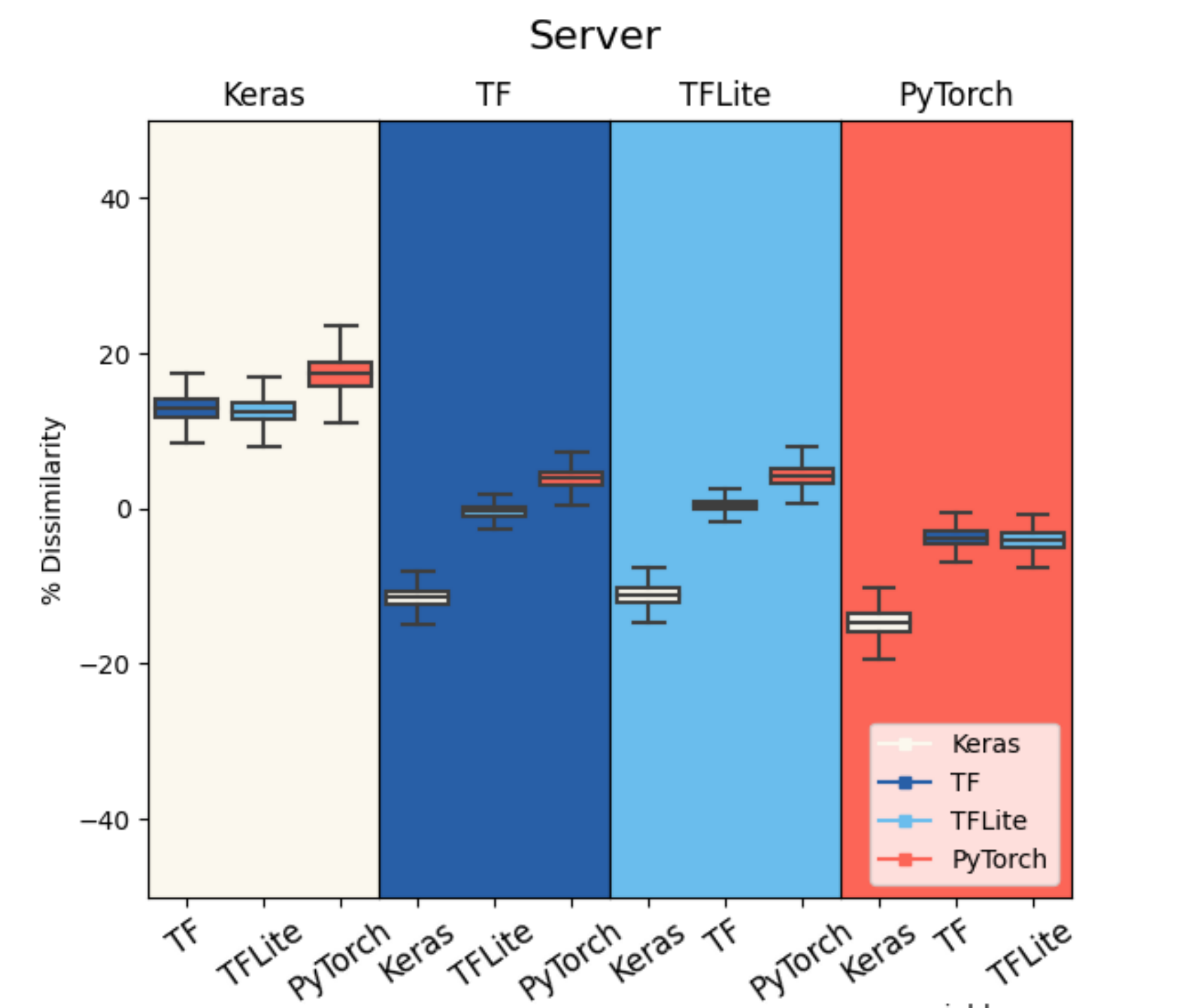}
	\caption{Comparison of inference times (\%) across DL Frameworks for \texttt{MobileNetV2} on Server with \textit{Default} optimization.}
	\label{fig:devices-change}
	\vspace{-10pt}
\end{figure}

We fixed the optimization setting to \texttt{Default} and device to \texttt{Server} and examined deviation in model inference times across models sourced from different DL frameworks. 
Results for MobileNetV2 are presented in Figure~\ref{fig:devices-change}.

It is worth noting that we observe considerable differences in inference times across model configurations. 
The extent of difference depends on the model. 
We find MobileNetV2 is most sensitive to DL framework changes (similar to output label), with inference times varying by $4 - 16$\%.
InceptionV3 was most robust to DL framework changes with an average difference of 8\% for changes between PyTorch and TF/TFLite. 
Finally, it is worth noting that a large difference  of upto $5\times$ was observed onResNet101 between PyTorch and other frameworks. 
This is because ResNet101 uses PyTorch version1, unlike other frameworks using version 2.


\begin{figure}[!t]
	\centering
	\includegraphics[width=0.7\columnwidth]{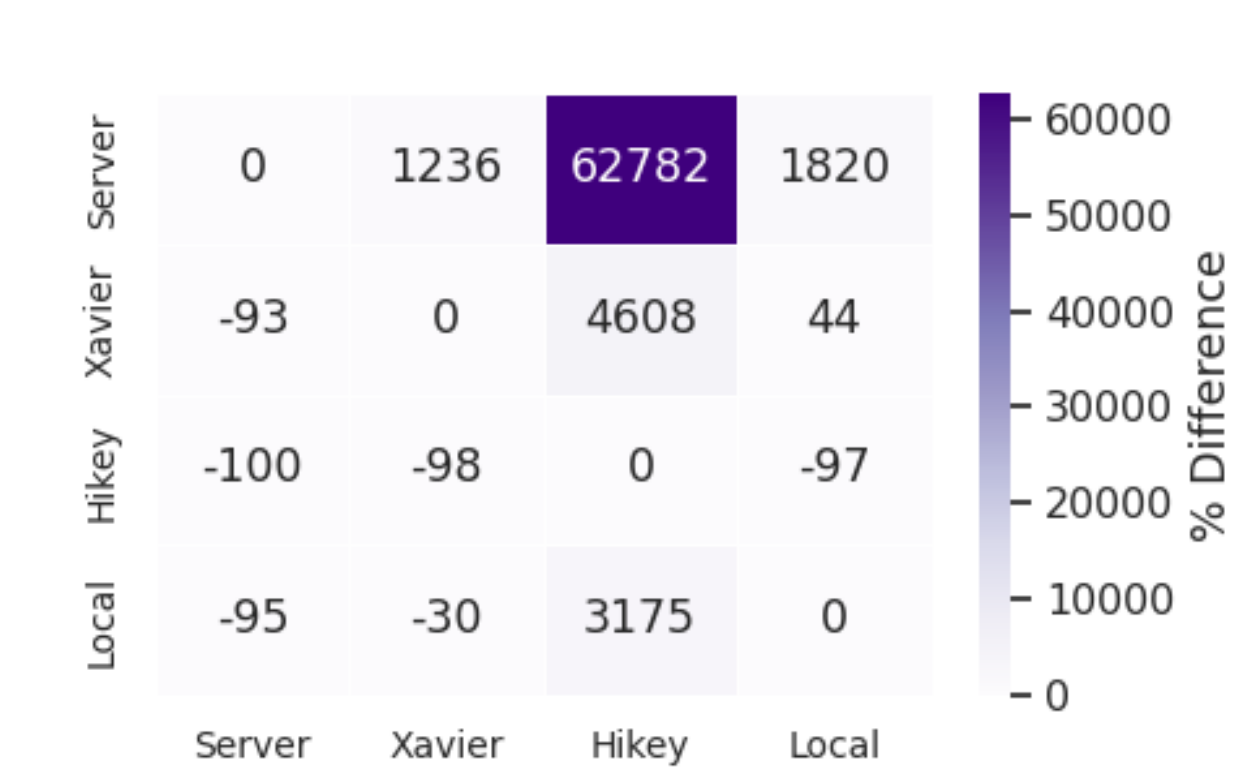}
	\caption{Device execution times difference (\%) on \textit{InceptionV3, TensorFlow, Default Optimization}.}
	\label{fig:devices-change-inception}
\end{figure}


\subsubsection{Varying Compiler Optimization}

With device fixed to \texttt{Server} and framework to Keras, we vary optimization levels between \emph{Basic}, \emph{Default}, and \emph{Extended}, examining the effect on model inference times.
We find changing optimization levels has a sizeable effect on inference times. 

Execution times generally improved with increased optimization as we moved from \texttt{Basic} to \texttt{Extended}, up to  $121.4\%$ across optimizations in all models. There are, however, some exceptions. For instance, MobileNetV2 using PyTorch was 11\% slower on \texttt{Hikey} device when using \texttt{Default} versus the \texttt{Basic} optimization. 
This suggests that increased optimization does not always result in speedup and in some cases can degrade performance. 


\subsubsection{Varying Hardware Accelerator}

We measured model inference on different hardware devices while fixing the DL framework and optimization level.
Figure~\ref{fig:devices-change-inception} shows the results when evaluating \textit{InceptionV3} model sourced from \textit{TensorFlow} with \textit{Default} optimization. 
We compute average inference time across all images in the test set and use this in comparing between devices. 
As expected, model inference times vary considerably with devices, based on their processing power and memory. 
Maximum execution time difference was observed between \texttt{Server} and \texttt{Hikey} (62782\%) on InceptionV3 and the minimum difference was 41\%, between \texttt{Xavier} and ~\texttt{Local} devices in ResNet101V2, utilizing \textit{Default} optimization.
Inference time difference with device change is most significant on InceptionV3 (upto $627\times$) owing to its larger size and memory requirements. When running on low-end device like Hikey as opposed to Server, we believe the smaller memory can cause cache misses resulting in significant inference time penalty.


\section{Threats To Validity}

There are five threats to validity in our experiments based on the dataset, models, model pre-processing, compiler framework and inference time.

First, we only evaluate robustness using four image recognition models that are widely used.
Results are model dependent as seen in our experiments and will likely vary on other models.  
We plan on conducting a more extensive evaluation in the future.

Second, we use the ImageNet~\cite{ILSVRC17} image classification test dataset for our experiments. 
This widely used dataset is a common benchmark which we believe adequately stresses the different model configurations in terms of both output label and inference time. 
Other datasets may yield different robustness results on the models considered. 

Third, model pre-processing is crucial for model performance~\cite{preprocessingimportance}.
In our experiments, we use the recommended pre-processing for each model and DL framework. Results may vary for other pre-processing settings. 
 
The fourth threat is introduced by the use of the TVM compiler framework and importing models into it.
To ensure that errors are not introduced in this process, for each model we validated the output labels for 100 random image samples from their source framework against the output label given by the model after importing into TVM.

The final threat is in inference time measurement.
To ensure time deviations (especially in the first or ``cold'' run) are taken into account, we repeat inferences for each image 10 times and use the average inference time across 10 runs.

\section{Discussion}

We make the following observations based on our results:

\textbf{Sensitivity to DL framework} 
Changing the DL framework used to generate the model can have a considerable effect on both the output label, and surprisingly, the model inference time.
The extent of this impact depends on other environment parameters.
Among the four models, MobileNetV2 had $57\%$ dissimilarity in output labels when PyTorch source is changed to TF/TFLite.  
Other models also had significant sensitivity to DL framework, up to $57\%$.

Inference time impact varied from $1-16\%$ (not including ResNet101 PyTorch version).

\textbf{Sensitivity to Framework Conversion}
Framework conversion between TF or PyTorch to TFLite can have an effect on output labels for some models.
ResNet101V2 was most sensitive to conversion from PyTorch to TFLite with 37\% disssimilarity in output labels. 
InceptionV3 was the most robust to framework conversion with respect to output labels, with a change between 0 and 4\%.

\textbf{Sensitivity to Compiler Optimization}
Compiler optimizations had no effect on output labels, but as expected have a considerable effect on model inference time, ranging from $1\%$ to $16\%$ across models.
This is not surprising as different optimization settings have a direct effect on code efficiency.

\textbf{Sensitivity to Hardware Devices}
Changing hardware devices had no impact on output labels from all four models but had a considerable effect on inference time, ranging from $1.2$ to $628\times$ and owing to the wide range in device capabilities used in our experiments. The device change impact was most seen on InceptionV3, with \textit{Default} optimization. For instance, when we change from Server to Hikey device, we find inference time is $87\times$ slower on MobileNetV2, $227\times$ slower on ResNet101, $285\times$ slower on DenseNet121 and $628\times$ slower on InceptionV3. 

In safety-critical applications, the consequences of the above sensitivities can be crucial.
Therefore it is essential that framework, compiler, and hardware communities, along with the developers of DNN models are aware of these sources of error, and test their systems for robustness to computational environment changes.
Currently, there is no regulation or benchmarking of DNN model performance for environment parameter configurations. 
The results from our experiments indicate that assessing sensitivity to environment parameters is an important consideration during model development and use.


\bibliography{References}

\begin{thebibliography}{38}
\providecommand{\natexlab}[1]{#1}

\bibitem[{Com(2022)}]{ComputeLibrary2022}
 2022.
\newblock Compute {{Library}}.
\newblock Arm Software.

\bibitem[{onn(2022)}]{onnx}
 2022.
\newblock Open Neural Network Exchange.
\newblock https://onnx.ai/.

\bibitem[{Abadi et~al.(2015)Abadi, Agarwal, Barham, Brevdo, Chen, Citro,
  Corrado, Davis, Dean, Devin, Ghemawat, Goodfellow, Harp, Irving, Isard, Jia,
  Jozefowicz, Kaiser, Kudlur, Levenberg, Man\'{e}, Monga, Moore, Murray, Olah,
  Schuster, Shlens, Steiner, Sutskever, Talwar, Tucker, Vanhoucke, Vasudevan,
  Vi\'{e}gas, Vinyals, Warden, Wattenberg, Wicke, Yu, and
  Zheng}]{tensorflow2015-whitepaper}
Abadi, M.; Agarwal, A.; Barham, P.; Brevdo, E.; Chen, Z.; Citro, C.; Corrado,
  G.~S.; Davis, A.; Dean, J.; Devin, M.; Ghemawat, S.; Goodfellow, I.; Harp,
  A.; Irving, G.; Isard, M.; Jia, Y.; Jozefowicz, R.; Kaiser, L.; Kudlur, M.;
  Levenberg, J.; Man\'{e}, D.; Monga, R.; Moore, S.; Murray, D.; Olah, C.;
  Schuster, M.; Shlens, J.; Steiner, B.; Sutskever, I.; Talwar, K.; Tucker, P.;
  Vanhoucke, V.; Vasudevan, V.; Vi\'{e}gas, F.; Vinyals, O.; Warden, P.;
  Wattenberg, M.; Wicke, M.; Yu, Y.; and Zheng, X. 2015.
\newblock {TensorFlow}: Large-Scale Machine Learning on Heterogeneous Systems.
\newblock Software available from tensorflow.org.

\bibitem[{Camacho{-}Collados and Pilehvar(2017)}]{preprocessingimportance}
Camacho{-}Collados, J.; and Pilehvar, M.~T. 2017.
\newblock On the Role of Text Preprocessing in Neural Network Architectures: An
  Evaluation Study on Text Categorization and Sentiment Analysis.
\newblock \emph{CoRR}, abs/1707.01780.

\bibitem[{Chen et~al.(2018{\natexlab{a}})Chen, Moreau, Jiang, Zheng, Yan, Shen,
  Cowan, Wang, Hu, Ceze, Guestrin, and Krishnamurthy}]{tvm}
Chen, T.; Moreau, T.; Jiang, Z.; Zheng, L.; Yan, E.; Shen, H.; Cowan, M.; Wang,
  L.; Hu, Y.; Ceze, L.; Guestrin, C.; and Krishnamurthy, A. 2018{\natexlab{a}}.
\newblock {{TVM}}: {{An}} Automated End-to-End Optimizing Compiler for Deep
  Learning.
\newblock In \emph{13th {{USENIX}} Symposium on Operating Systems Design and
  Implementation ({{OSDI}} 18)}, 578--594.
\newblock ISBN 978-1-939133-08-3.

\bibitem[{Chen et~al.(2018{\natexlab{b}})Chen, Zheng, Yan, Jiang, Moreau, Ceze,
  Guestrin, and Krishnamurthy}]{chenLearningOptimizeTensor2018}
Chen, T.; Zheng, L.; Yan, E.; Jiang, Z.; Moreau, T.; Ceze, L.; Guestrin, C.;
  and Krishnamurthy, A. 2018{\natexlab{b}}.
\newblock Learning to {{Optimize Tensor Programs}}.
\newblock In Bengio, S.; Wallach, H.; Larochelle, H.; Grauman, K.;
  {Cesa-Bianchi}, N.; and Garnett, R., eds., \emph{Advances in {{Neural
  Information Processing Systems}} 31}, 3393--3404. {Curran Associates, Inc.}

\bibitem[{Chetlur et~al.(2014)Chetlur, Woolley, Vandermersch, Cohen, Tran,
  Catanzaro, and Shelhamer}]{chetlur2014cudnn}
Chetlur, S.; Woolley, C.; Vandermersch, P.; Cohen, J.; Tran, J.; Catanzaro, B.;
  and Shelhamer, E. 2014.
\newblock Cudnn: {{Efficient}} Primitives for Deep Learning.
\newblock \emph{arXiv preprint arXiv:1410.0759}.

\bibitem[{Chiu et~al.(2020)Chiu, Tsai, Ruan, Shen, and
  Lee}]{chiuMobilenetSSDv2ImprovedObject2020}
Chiu, Y.-C.; Tsai, C.-Y.; Ruan, M.-D.; Shen, G.-Y.; and Lee, T.-T. 2020.
\newblock Mobilenet-{{SSDv2}}: {{An Improved Object Detection Model}} for
  {{Embedded Systems}}.
\newblock In \emph{2020 {{International Conference}} on {{System Science}} and
  {{Engineering}} ({{ICSSE}})}, 1--5.

\bibitem[{Chollet et~al.(2015)}]{chollet2015keras}
Chollet, F.; et~al. 2015.
\newblock Keras.
\newblock \url{https://keras.io}.

\bibitem[{Dai et~al.(2021)Dai, Liu, Le, and
  Tan}]{daiCoAtNetMarryingConvolution2021}
Dai, Z.; Liu, H.; Le, Q.~V.; and Tan, M. 2021.
\newblock {{CoAtNet}}: {{Marrying Convolution}} and {{Attention}} for {{All
  Data Sizes}}.
\newblock In \emph{Advances in {{Neural Information Processing Systems}}},
  volume~34, 3965--3977. {Curran Associates, Inc.}

\bibitem[{Dreossi et~al.(2019)Dreossi, Fremont, Ghosh, Kim, Ravanbakhsh,
  Vazquez-Chanlatte, and Seshia}]{dreossi2019verifai}
Dreossi, T.; Fremont, D.~J.; Ghosh, S.; Kim, E.; Ravanbakhsh, H.;
  Vazquez-Chanlatte, M.; and Seshia, S.~A. 2019.
\newblock VERIFAI: A toolkit for the design and analysis of artificial
  intelligence-based systems.
\newblock \emph{arXiv preprint arXiv:1902.04245}.

\bibitem[{Guo et~al.(2021)Guo, Zhao, Song, and
  Jiang}]{guoCoverageGuidedDifferential2021}
Guo, J.; Zhao, Y.; Song, H.; and Jiang, Y. 2021.
\newblock Coverage {{Guided Differential Adversarial Testing}} of {{Deep
  Learning Systems}}.
\newblock \emph{IEEE Transactions on Network Science and Engineering}, 8(2):
  933--942.

\bibitem[{He et~al.(2015)He, Zhang, Ren, and Sun}]{resnet}
He, K.; Zhang, X.; Ren, S.; and Sun, J. 2015.
\newblock Deep Residual Learning for Image Recognition.
\newblock \emph{CoRR}, abs/1512.03385.

\bibitem[{Huang, Liu, and Weinberger(2016)}]{densenet}
Huang, G.; Liu, Z.; and Weinberger, K.~Q. 2016.
\newblock Densely Connected Convolutional Networks.
\newblock \emph{CoRR}, abs/1608.06993.

\bibitem[{Humbatova et~al.(2020)Humbatova, Jahangirova, Bavota, Riccio, Stocco,
  and Tonella}]{humbatova2020taxonomy}
Humbatova, N.; Jahangirova, G.; Bavota, G.; Riccio, V.; Stocco, A.; and
  Tonella, P. 2020.
\newblock Taxonomy of real faults in deep learning systems.
\newblock In \emph{Proceedings of the ACM/IEEE 42nd International Conference on
  Software Engineering}, 1110--1121.

\bibitem[{Khan et~al.(2019)Khan, Sohail, Zahoora, and Qureshi}]{SurveyCNNs}
Khan, A.; Sohail, A.; Zahoora, U.; and Qureshi, A.~S. 2019.
\newblock A Survey of the Recent Architectures of Deep Convolutional Neural
  Networks.
\newblock \emph{CoRR}, abs/1901.06032.

\bibitem[{Liu et~al.(2018)Liu, Wu, Wei, Cao, Sahin, and
  Zhang}]{benchmarkingdlframeworkmetrics}
Liu, L.; Wu, Y.; Wei, W.; Cao, W.; Sahin, S.; and Zhang, Q. 2018.
\newblock Benchmarking Deep Learning Frameworks: Design Considerations, Metrics
  and Beyond.
\newblock In \emph{2018 IEEE 38th International Conference on Distributed
  Computing Systems (ICDCS)}, 1258--1269.

\bibitem[{Liu et~al.(2020)Liu, Chen, Zhang, Qin, Ji, Lin, and
  Yang}]{liuEnhancingInteroperabilityDeep2020}
Liu, Y.; Chen, C.; Zhang, R.; Qin, T.; Ji, X.; Lin, H.; and Yang, M. 2020.
\newblock Enhancing the Interoperability between Deep Learning Frameworks by
  Model Conversion.
\newblock In \emph{Proceedings of the 28th {{ACM Joint Meeting}} on {{European
  Software Engineering Conference}} and {{Symposium}} on the {{Foundations}} of
  {{Software Engineering}}}, {{ESEC}}/{{FSE}} 2020, 1320--1330. {New York, NY,
  USA}: {Association for Computing Machinery}.
\newblock ISBN 978-1-4503-7043-1.

\bibitem[{Mahmoud et~al.(2019)Mahmoud, Essam, Elshawi, and Sakr}]{dlbench}
Mahmoud, N.; Essam, Y.; Elshawi, R.; and Sakr, S. 2019.
\newblock DLBench: An Experimental Evaluation of Deep Learning Frameworks.
\newblock In \emph{2019 IEEE International Congress on Big Data
  (BigDataCongress)}, 149--156.

\bibitem[{Paszke et~al.(2019)Paszke, Gross, Massa, Lerer, Bradbury, Chanan,
  Killeen, Lin, Gimelshein, Antiga, Desmaison, K{\"{o}}pf, Yang, DeVito,
  Raison, Tejani, Chilamkurthy, Steiner, Fang, Bai, and Chintala}]{pytorch}
Paszke, A.; Gross, S.; Massa, F.; Lerer, A.; Bradbury, J.; Chanan, G.; Killeen,
  T.; Lin, Z.; Gimelshein, N.; Antiga, L.; Desmaison, A.; K{\"{o}}pf, A.; Yang,
  E.~Z.; DeVito, Z.; Raison, M.; Tejani, A.; Chilamkurthy, S.; Steiner, B.;
  Fang, L.; Bai, J.; and Chintala, S. 2019.
\newblock PyTorch: An Imperative Style, High-Performance Deep Learning Library.
\newblock \emph{CoRR}, abs/1912.01703.

\bibitem[{Pei et~al.(2017)Pei, Cao, Yang, and Jana}]{deepxplore}
Pei, K.; Cao, Y.; Yang, J.; and Jana, S. 2017.
\newblock DeepXplore: Automated Whitebox Testing of Deep Learning Systems.
\newblock \emph{CoRR}, abs/1705.06640.

\bibitem[{Pham et~al.(2019)Pham, Lutellier, Qi, and Tan}]{pham2019cradle}
Pham, H.~V.; Lutellier, T.; Qi, W.; and Tan, L. 2019.
\newblock CRADLE: Cross-Backend Validation to Detect and Localize Bugs in Deep
  Learning Libraries.
\newblock In \emph{2019 IEEE/ACM 41st International Conference on Software
  Engineering (ICSE)}, 1027--1038.

\bibitem[{Russakovsky et~al.(2015)Russakovsky, Deng, Su, Krause, Satheesh, Ma,
  Huang, Karpathy, Khosla, Bernstein, Berg, and Fei-Fei}]{ILSVRC17}
Russakovsky, O.; Deng, J.; Su, H.; Krause, J.; Satheesh, S.; Ma, S.; Huang, Z.;
  Karpathy, A.; Khosla, A.; Bernstein, M.; Berg, A.~C.; and Fei-Fei, L. 2015.
\newblock {ImageNet Large Scale Visual Recognition Challenge}.
\newblock \emph{International Journal of Computer Vision (IJCV)}, 115(3):
  211--252.

\bibitem[{Sandler et~al.(2018)Sandler, Howard, Zhu, Zhmoginov, and
  Chen}]{mobilenetv2}
Sandler, M.; Howard, A.~G.; Zhu, M.; Zhmoginov, A.; and Chen, L. 2018.
\newblock Inverted Residuals and Linear Bottlenecks: Mobile Networks for
  Classification, Detection and Segmentation.
\newblock \emph{CoRR}, abs/1801.04381.

\bibitem[{Shen et~al.(2021)Shen, Ma, Chen, Tian, Cheung, and
  Chen}]{dlcompileroptimisationbugs}
Shen, Q.; Ma, H.; Chen, J.; Tian, Y.; Cheung, S.-C.; and Chen, X. 2021.
\newblock A Comprehensive Study of Deep Learning Compiler Bugs.
\newblock In \emph{Proceedings of the 29th ACM Joint Meeting on European
  Software Engineering Conference and Symposium on the Foundations of Software
  Engineering}, ESEC/FSE 2021, 968–980. New York, NY, USA: Association for
  Computing Machinery.
\newblock ISBN 9781450385626.

\bibitem[{Shi et~al.(2016)Shi, Wang, Xu, and Chu}]{benchmarkdlswtools}
Shi, S.; Wang, Q.; Xu, P.; and Chu, X. 2016.
\newblock Benchmarking State-of-the-Art Deep Learning Software Tools.
\newblock \emph{CoRR}, abs/1608.07249.

\bibitem[{Shorten and Khoshgoftaar(2019)}]{shorten2019survey}
Shorten, C.; and Khoshgoftaar, T.~M. 2019.
\newblock A survey on image data augmentation for deep learning.
\newblock \emph{Journal of big data}, 6(1): 1--48.

\bibitem[{Szegedy et~al.(2015)Szegedy, Vanhoucke, Ioffe, Shlens, and
  Wojna}]{inceptionv3}
Szegedy, C.; Vanhoucke, V.; Ioffe, S.; Shlens, J.; and Wojna, Z. 2015.
\newblock Rethinking the Inception Architecture for Computer Vision.
\newblock \emph{CoRR}, abs/1512.00567.

\bibitem[{Tian et~al.(2018)Tian, Pei, Jana, and Ray}]{tian2018deeptest}
Tian, Y.; Pei, K.; Jana, S.; and Ray, B. 2018.
\newblock Deeptest: Automated testing of deep-neural-network-driven autonomous
  cars.
\newblock In \emph{Proceedings of the 40th international conference on software
  engineering}, 303--314.

\bibitem[{Turner et~al.(2018)Turner, Cano, Radu, Crowley, O’Boyle, and
  Storkey}]{iiswc_2018}
Turner, J.; Cano, J.; Radu, V.; Crowley, E.~J.; O’Boyle, M.; and Storkey, A.
  2018.
\newblock {Characterising Across-Stack Optimisations for Deep Convolutional
  Neural Networks}.
\newblock In \emph{IISWC}, 101--110.

\bibitem[{Vaswani et~al.(2017)Vaswani, Shazeer, Parmar, Uszkoreit, Jones,
  Gomez, Kaiser, and Polosukhin}]{vaswani2017attention}
Vaswani, A.; Shazeer, N.; Parmar, N.; Uszkoreit, J.; Jones, L.; Gomez, A.~N.;
  Kaiser, {\L}.; and Polosukhin, I. 2017.
\newblock Attention Is {{All You Need}}.
\newblock In Guyon, I.; Luxburg, U.~V.; Bengio, S.; Wallach, H.; Fergus, R.;
  Vishwanathan, S.; and Garnett, R., eds., \emph{Advances in {{Neural
  Information Processing Systems}} 30}, 5998--6008. {Curran Associates, Inc.}

\bibitem[{Wang et~al.(2020)Wang, Yan, Chen, Liu, and Zhang}]{wang2020lemon}
Wang, Z.; Yan, M.; Chen, J.; Liu, S.; and Zhang, D. 2020.
\newblock Deep Learning Library Testing via Effective Model Generation.
\newblock In \emph{Proceedings of the 28th ACM Joint Meeting on European
  Software Engineering Conference and Symposium on the Foundations of Software
  Engineering}, ESEC/FSE 2020, 788–799. New York, NY, USA: Association for
  Computing Machinery.
\newblock ISBN 9781450370431.

\bibitem[{Wu et~al.(2022)Wu, Liu, Pu, Cao, Sahin, Wei, and
  Zhang}]{studydlframeworks}
Wu, Y.; Liu, L.; Pu, C.; Cao, W.; Sahin, S.; Wei, W.; and Zhang, Q. 2022.
\newblock A Comparative Measurement Study of Deep Learning as a Service
  Framework.
\newblock \emph{IEEE Transactions on Services Computing}, 15(1): 551--566.

\bibitem[{Xie et~al.(2019)Xie, Ma, Juefei-Xu, Xue, Chen, Liu, Zhao, Li, Yin,
  and See}]{xie2019deephunter}
Xie, X.; Ma, L.; Juefei-Xu, F.; Xue, M.; Chen, H.; Liu, Y.; Zhao, J.; Li, B.;
  Yin, J.; and See, S. 2019.
\newblock DeepHunter: A Coverage-Guided Fuzz Testing Framework for Deep Neural
  Networks.
\newblock In \emph{Proceedings of the 28th ACM SIGSOFT International Symposium
  on Software Testing and Analysis}, ISSTA 2019, 146–157. New York, NY, USA:
  Association for Computing Machinery.
\newblock ISBN 9781450362245.

\bibitem[{Zhai et~al.(2022)Zhai, Kolesnikov, Houlsby, and
  Beyer}]{zhaiScalingVisionTransformers2022}
Zhai, X.; Kolesnikov, A.; Houlsby, N.; and Beyer, L. 2022.
\newblock Scaling {{Vision Transformers}}.
\newblock In \emph{Proceedings of the {{IEEE}}/{{CVF Conference}} on {{Computer
  Vision}} and {{Pattern Recognition}}}, 12104--12113.

\bibitem[{Zhang et~al.(2018)Zhang, Zhang, Zhang, Liu, and
  Khurshid}]{zhang2018deeproad}
Zhang, M.; Zhang, Y.; Zhang, L.; Liu, C.; and Khurshid, S. 2018.
\newblock DeepRoad: GAN-based metamorphic testing and input validation
  framework for autonomous driving systems.
\newblock In \emph{2018 33rd IEEE/ACM International Conference on Automated
  Software Engineering (ASE)}, 132--142. IEEE.

\bibitem[{Zheng et~al.(2020)Zheng, Jia, Sun, Wu, Yu, {Haj-Ali}, Wang, Yang,
  Zhuo, Sen, Gonzalez, and Stoica}]{zhengAnsorGeneratingHighPerformance2020}
Zheng, L.; Jia, C.; Sun, M.; Wu, Z.; Yu, C.~H.; {Haj-Ali}, A.; Wang, Y.; Yang,
  J.; Zhuo, D.; Sen, K.; Gonzalez, J.~E.; and Stoica, I. 2020.
\newblock Ansor: {{Generating High-Performance Tensor Programs}} for {{Deep
  Learning}}.
\newblock In \emph{14th {{USENIX Symposium}} on {{Operating Systems Design}}
  and {{Implementation}} ({{OSDI}} 20)}, 863--879.
\newblock ISBN 978-1-939133-19-9.

\bibitem[{Zhou et~al.(2020)Zhou, Li, Kong, Guo, Zhang, Yu, Zhang, and
  Liu}]{zhou2020deepbillboard}
Zhou, H.; Li, W.; Kong, Z.; Guo, J.; Zhang, Y.; Yu, B.; Zhang, L.; and Liu, C.
  2020.
\newblock DeepBillboard: Systematic Physical-World Testing of Autonomous
  Driving Systems.
\newblock In \emph{Proceedings of the ACM/IEEE 42nd International Conference on
  Software Engineering}, ICSE '20, 347–358. New York, NY, USA: Association
  for Computing Machinery.
\newblock ISBN 9781450371216.

\end{thebibliography}

\end{document}